\documentclass[preprint,12pt]{elsarticle}

\usepackage{amssymb}
\usepackage{amsmath}
\usepackage{tcolorbox}
\usepackage{booktabs}
\usepackage{subcaption}

\usepackage{amssymb}
\usepackage{amsmath}
\usepackage{tcolorbox}
\usepackage{booktabs}
\usepackage{subcaption}

\journal{arXiv}

\begin{document}

\begin{frontmatter}



\title{Theia: Large-Scale Multimodal Captioning and Automated Validation of the Incidents1M Dataset for Data-Free Distillation} 


\author[1]{Simone Giano} 
\author[1]{Lorenzo Severini} 
\author[1]{Alessandro Galdelli} 
\author[1]{Adriano Mancini} 

\affiliation[1]{organization={Dipartimento di Ingegneria dell'Informazione, Università Politecnica delle Marche},
            addressline={Via Brecce Bianche 12}, 
            city={Ancona},
            postcode={60121}, 
            state={Marche},
            country={Italy}}

\begin{abstract}
The deployment of Vision-Language Models (VLMs) in critical domains like disaster management requires high-quality multimodal datasets, especially for transferring knowledge via Data-Free Knowledge Distillation (DFKD). However, existing datasets in this domain either entirely lack descriptive text, such as Incidents1M, or suffer from severe text-image semantic misalignment, such as CrisisMMD. In this work, we present a novel methodology to construct and automatically validate a large-scale multimodal dataset for disaster response. Starting from the vision-only Incidents1M, we successfully recovered 100,000 images and generated high-fidelity textual descriptions using two distinct Qwen3.5 architectures: a 4B dense model and a 35B Mixture-of-Experts (MoE) model. To ensure the generated captions provide reliable semantic anchoring for DFKD, we introduce an image-blind LLM-as-a-Judge validation pipeline leveraging Qwen3.5-9B. By intentionally obscuring the original image from the judge, this evaluator accurately simulates the modality gap of the student model during data-free distillation. Our evaluation across 173,179 label pairs demonstrates a high semantic agreement (78.65/100) between the two architectures. Furthermore, the automated evaluation reveals a conservative captioning behaviour, characterized by a high Precision (77.6\%) and low Recall (46.0\%). This minimizes the false positive noise, while simultaneously exposing underlying human annotation inconsistencies in the original ground truth. This work provides a scalable, LLM-validated multimodal dataset and a reproducible framework to advance cross-modal knowledge distillation. 
\end{abstract}



\end{frontmatter}

\section{Introduction}
\label{sec:introduction}

The contemporary landscape of Artificial Intelligence has been profoundly reshaped by the transition from specialized unimodal systems to unified Vision-Language Models (VLMs) \cite{Scaling_visual, Flamingo}. By projecting visual and textual inputs into a shared semantic space, foundational VLMs are now capable of perceiving, understanding, and reasoning simultaneously over heterogeneous data streams \cite{BLIP2, Visual_instruction_tuning}. This architectural convergence has unlocked unprecedented zero-shot capabilities across diverse fields, ranging from multimodal search engines \cite{Amazon_Shop, Learning_transfer} to autonomous robotics \cite{RT2, PALME}, medical diagnostics \cite{Generalisti_bio_AI, luVisuallanguageFoundationModel2024a}, and digital accessibility \cite{VizWiz, Caption_blind}.

Despite these theoretical advancements, the operational efficacy of VLMs remains strictly bound to the availability of high-quality multimodal data. This reliance becomes critical in sensitive and high-stakes domains, such as disaster management and emergency response \cite{multimodalbaseline2020}. Recognizing a disaster scenario, distinguishing a flooded area from a normal coastline or a collapsed building from a construction site, requires complex contextual reasoning \cite{Detecting_natural_disaster}. However, existing datasets in this domain suffer from significant structural limitations. For instance, while CrisisMMD \cite{CrisisMMD} provides genuine social media imagery, its textual annotations are often noisy, sarcastic, or entirely disconnected from the visual content. On the contrary, the Incidents1M dataset \cite{Incidents1M} offers a massive, rigorously verified collection of over a million images depicting 43 disaster categories, but lacks entirely textual descriptions.

The absence of dense, explanatory textual annotations poses a severe bottleneck for transferring knowledge from massive cloud-based VLMs \cite{QWEN-VL, qwen2025qwen25technicalreport} to compact models suitable for edge computing or drones. This transfer is typically achieved via Knowledge Distillation \cite{HintonKD, SurveyKD}, and specifically through Data-Free Knowledge Distillation (DFKD) \cite{DFLSN, DeepInversion, FastDKFD} when original training images are unavailable due to privacy or licensing constraints. In a multimodal DFKD scenario, the textual modality acts as the sole semantic anchor to synthesize surrogate visual data. Without highly accurate and objective captions, the generative process is compromised, forcing the student model to inherit flawed representations. Furthermore, recent studies highlight the propensity of VLMs to suffer from object hallucinations \cite{li-etal-2023-evaluating}, making the automated generation of these anchors a delicate task.

To bridge this gap, we present a large-scale multimodal extension of the Incidents1M dataset. Starting from the original vision-only ground truth, we successfully recovered 100,000 images and enriched them with 200,000 high-fidelity captions. To evaluate the impact of different architectural paradigms on the captioning task, we employed two distinct implementations of the state-of-the-art open-source Qwen3.5 family: a dense 4B parameter model and a 35B parameter Mixture-of-Experts (MoE) model. The MoE paradigm allows accessing a vastly larger parametric knowledge base while maintaining competitive inference times through sparse routing \cite{Sparsly_gated_moe, jiang2024mixtralexperts}.

Finally, to rigorously validate the generated text without resorting to unscalable human annotation or rigid syntactic metrics, we implemented an automated evaluation pipeline leveraging the LLM-as-a-Judge paradigm \cite{MTBench, GEval, kim2024prometheus}. Crucially, our evaluator (Qwen3.5-9B) is intentionally kept image-blind. By forcing the judge to extract disaster labels relying solely on the generated text, we accurately simulate the modality gap that a student model faces during Data-Free Distillation. 

This work makes the following main contributions:
\begin{itemize}
    \item We provide a massive multimodal disaster dataset comprising 100,000 images and 200,000 descriptive captions, bridging the critical gap between visual incidents data and semantic text.
    \item We present a comparative analysis of captioning behaviour between dense and Mixture-of-Experts (MoE) VLM architectures in a continuous batching inference scenario.
    \item We introduce an image-blind LLM-as-a-Judge validation framework that quantifies the transferability of essential visual information into text, demonstrating a conservative, high-precision generative behaviour suitable for DFKD.
    \item We expose hidden human-annotation inconsistencies within the original Incidents1M ground truth, demonstrating the efficacy of modern VLMs in auditing existing large-scale benchmarks.
\end{itemize}

\section{Related Work}
\label{sec:related_work}

\subsection{Vision-Language Models and Mixture-of-Experts}
The evolution of multimodal artificial intelligence has been driven by the development of Vision-Language Models (VLMs) capable of aligning visual representations with textual semantics \cite{Scaling_visual}. Pioneering architectures like Flamingo \cite{Flamingo} and BLIP-2 \cite{BLIP2} demonstrated the effectiveness of bridging frozen image encoders with Large Language Models (LLMs) to achieve powerful zero-shot generalization. This paradigm was further standardized by visual instruction tuning techniques, notably LLaVA \cite{Visual_instruction_tuning}, which empowered models to follow complex multimodal human instructions. As VLMs scaled, the computational bottleneck was mitigated by the adoption of sparse architectures, specifically the Mixture-of-Experts (MoE) paradigm \cite{Sparsly_gated_moe}. By dynamically routing tokens to a subset of specialized neural networks, MoE models like Mixtral \cite{jiang2024mixtralexperts} achieve the representational capacity of massive dense models while maintaining highly efficient inference costs. In this work, we leverage the state-of-the-art Qwen family \cite{QWEN-VL, qwen2025qwen25technicalreport}, exploiting both its dense and MoE variations to generate high-fidelity annotations at scale.

\subsection{Datasets for Disaster Response}
Applying deep learning to disaster management requires robust datasets to train models capable of understanding chaotic and unstructured environments \cite{Detecting_natural_disaster}. Early efforts heavily relied on social media scraping. For instance, CrisisMMD \cite{CrisisMMD} and related baselines \cite{multimodalbaseline2020} provided multimodal datasets comprising images and textual tweets collected during natural disasters. However, the inherent noise of social media, where images are frequently paired with unrelated text, severely limits their utility for strict visual-semantic alignment. To overcome textual ambiguity, the Incidents1M dataset \cite{Incidents1M} introduced a massive, vision-only benchmark. It provides over a million rigorously verified images annotated with 43 multi-label incident categories, deliberately discarding the noisy textual component. While Incidents1M represents the gold standard for visual disaster classification, its strictly unimodal nature prevents its direct application in modern cross-modal training pipelines.

\subsection{Data-Free Knowledge Distillation}
Knowledge Distillation (KD) \cite{HintonKD} is a well-established technique to compress large, unwieldy models (Teachers) into compact, deployable ones (Students) \cite{SurveyKD}. However, traditional KD assumes full access to the original training data. When privacy constraints, copyright issues, or sheer dataset sizes make the original images unavailable, Data-Free Knowledge Distillation (DFKD) is used. Foundational DFKD approaches \cite{DFLSN, DeepInversion} invert the pre-trained Teacher to synthesize surrogate training data. Recent advancements have successfully accelerated this process \cite{FastDKFD}. In multimodal DFKD, the textual domain is strictly required to anchor the generation of surrogate images. If the text is missing or overly generic, the synthesized visual distribution collapses. Therefore, synthesizing a dense textual modality for established vision-only datasets like Incidents1M is a mandatory step to enable DFKD in critical scenarios.

\subsection{Automated Evaluation and LLM-as-a-Judge}
Evaluating the quality of generated multimodal text at an industrial scale poses a significant challenge. Traditional n-gram metrics (e.g., BLEU, ROUGE) exhibit severe limitations, as they penalize paraphrasing and fail to grasp logic negations. Moreover, VLMs are highly prone to object hallucinations, generating plausible but factually incorrect visual details \cite{li-etal-2023-evaluating}. To achieve scalable, semantic-aware validation, recent literature introduced the LLM-as-a-Judge paradigm \cite{MTBench}. Strong language models can be prompted to evaluate generated text against specific criteria with a high correlation to human judgment \cite{GEval}. Frameworks like Prometheus \cite{kim2024prometheus} demonstrated that providing an LLM with fine-grained evaluation rubrics yields highly reproducible assessments.

\subsection{Research Context and Our contribution}
Our research bridges the gaps identified across these domains. Unlike existing literature that either relies on noisy social media text \cite{CrisisMMD} or abandons text entirely \cite{Incidents1M}, we synthesize a clean multimodal dataset by pairing the rigorous visual taxonomy of Incidents1M with state-of-the-art VLM capabilities \cite{QWEN-VL}. Furthermore, we advance the LLM-as-a-Judge methodology \cite{MTBench, kim2024prometheus} by introducing an image-blind evaluation protocol. Instead of standard visual QA, our judge evaluates caption fidelity by verifying the persistence of ground-truth labels purely from text, thereby simulating the modality gap inherent to multimodal DFKD \cite{FastDKFD}. This ensures that our dataset is not merely descriptive, but strictly optimized for data-free knowledge transfer.

\section{Dataset Reconstruction and VLLM Pipeline}
\label{sec:pipeline}

To enable Data-Free Knowledge Distillation (DFKD), the visual domain must be paired with high-fidelity semantic annotations \cite{DFLSN}. Since the original Incidents1M dataset \cite{Incidents1M} is strictly vision-only, we engineered a comprehensive pipeline to physically reconstruct a subset of the dataset and generate the missing textual modality using state-of-the-art Vision-Language Models \cite{QWEN-VL}. 

\subsection{Mitigating Link Rot and Atomic Dataset Reconstruction}
The public release of Incidents1M does not distribute raw image files due to copyright and bandwidth constraints; instead, it provides a JSON file mapping multi-label annotations to the original hosting URLs \cite{Incidents1M}. Consequently, retrieving the dataset requires large-scale web scraping, a process severely hindered by link rot (e.g., 404 Not Found errors due to deleted content) and server-side bot protections. 

To overcome these challenges and secure a clean visual foundation, we developed a highly concurrent, fault-tolerant downloading orchestrator. To prevent the quiet accumulation of corrupted files, such as partial downloads or HTML error pages disguised as images, our system implements an atomic write strategy. Images are streamed in 64-kilobyte chunks into a temporary file (\texttt{.tmp}) and are renamed to their final \texttt{.jpg} extension only upon successful and complete byte verification. 

\begin{figure}[ht!]
    \centering
    \includegraphics[width=1\linewidth]{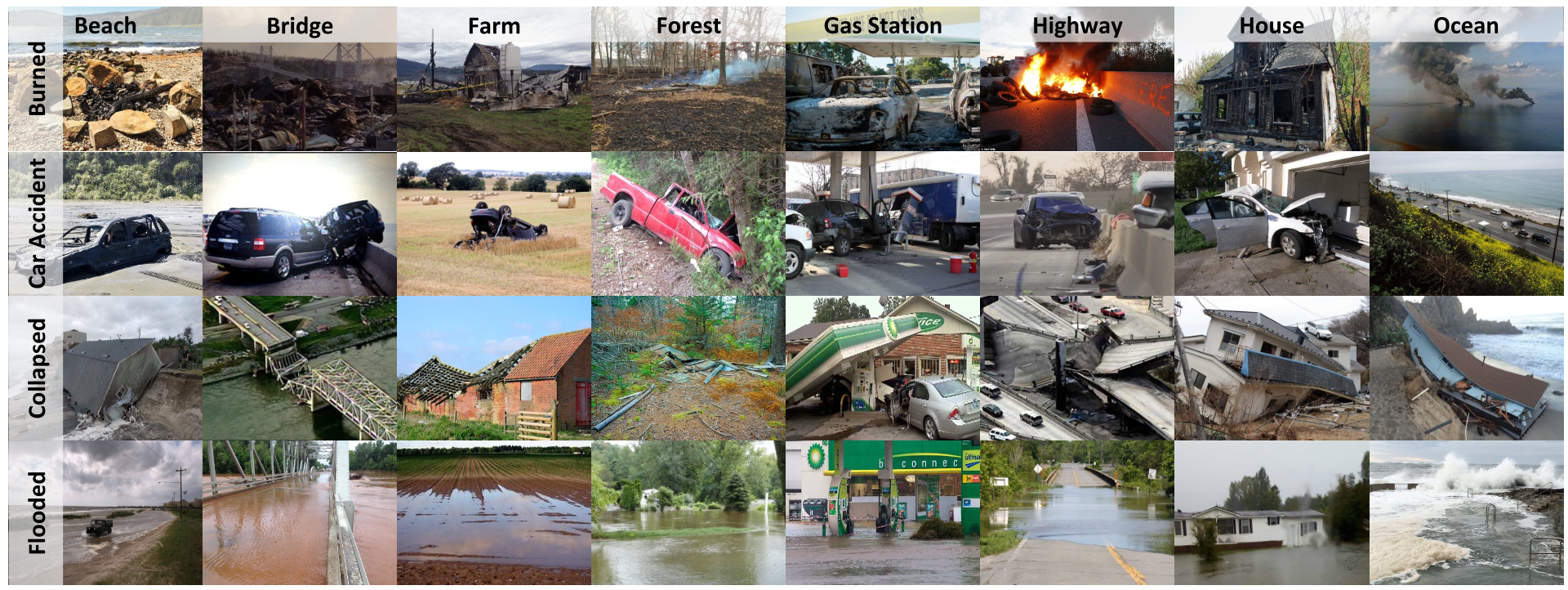}
    \caption{Image examples extracted from Incidents1M \cite{Incidents1M}.}
    \label{fig:IncidentsExample}
\end{figure}

Furthermore, simply processing the first 100,000 URLs would result in a heavily truncated dataset due to the unpredictable rate of dead links. To guarantee an exact number of 100,000 intact images, the orchestrator utilizes a dynamic reservation mechanism. Whenever a download irreversibly fails after exponential backoff retries, its reserved slot is released, and a new URL is automatically pulled from the remaining 1.7 million available links. This ensures the final dataset volume is precisely met with 100\% valid visual data, ready for multimodal inference (Fig. \ref{fig:IncidentsExample}).

\subsection{VLLM Architectures: Dense vs. Mixture-of-Experts}
For the generation of textual captions, we selected the Qwen3.5 vision-language family \cite{qwen2025qwen25technicalreport}, which natively supports early fusion of visual and textual tokens, projecting them into a shared representation space from the initial layers. To investigate how architectural scaling impacts the quality of semantic anchoring, we evaluated two distinct implementations on the same 100,000 images:
\begin{itemize}
    \item \textbf{Qwen3.5-4B (Dense):} A compact architecture featuring 4 billion parameters uniformly activated for every processed token. With 32 layers and a hidden dimension of 2560, it alternates linear attention (Gated DeltaNet) with traditional quadratic attention, optimizing latency and hardware footprint.
    \item \textbf{Qwen3.5-35B-A3B (Mixture-of-Experts):} A highly sparse architecture designed to expand the parametric knowledge base without a linear increase in computational cost. While it contains 35 billion parameters across 40 layers, its feed-forward networks act as routers, dynamically activating only 9 out of 256 experts (yielding approximately 3 billion active parameters) per token \cite{jiang2024mixtralexperts}.
\end{itemize}

\begin{figure}[htbp]
    \centering
    \includegraphics[width=0.75\linewidth]{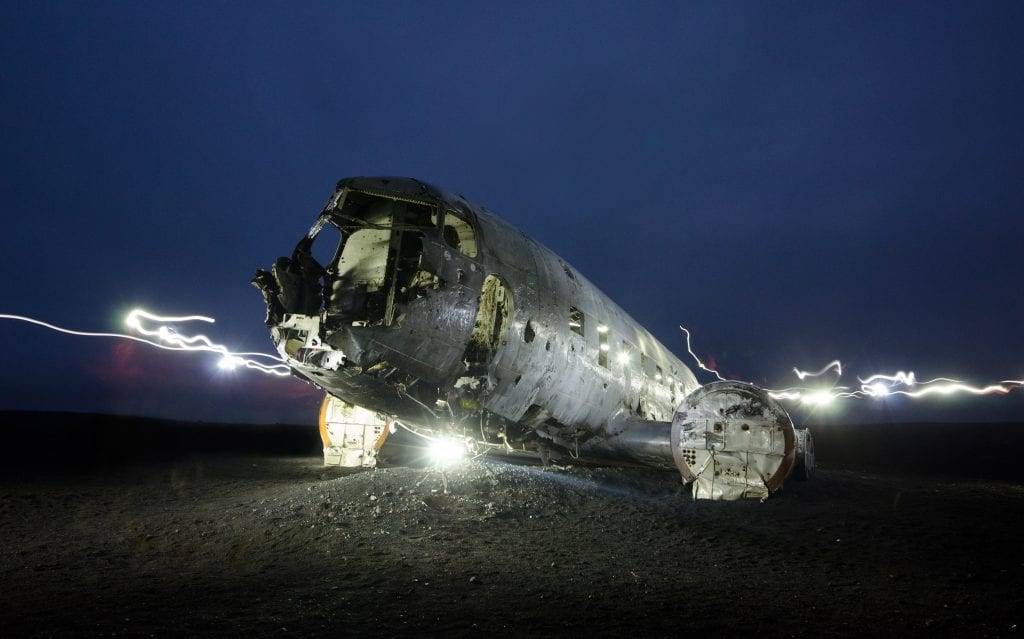}
    

    \begin{tcolorbox}[title={\footnotesize\itshape Generation Qwen3.5-4B}]
    \scriptsize
    "The image shows a pleasure boat, partially submerged in dark water, with the bow pointing to the left and the stern tilted to the right. The boat is light gray, with metallic details visible on the internal structure and outer edges. On the main deck, a small circular platform, probably an anchor windlass or a fastening device, is located in the center of the boat. On the right side, near the stern, there is a cylindrical object, similar to a tank or container, in off-white. The background is dominated by a dark blue night sky, with scattered clouds creating a dynamic effect. No other objects or subjects are present, and the scene is illuminated by artificial light coming from an unseen source, creating reflections on the water and the boat's surface. The composition is static, with the boat as the central and unique element, positioned slightly to the left of the center."
    \end{tcolorbox}

    \begin{tcolorbox}[title={\footnotesize\itshape Generation Qwen3.5-35B-A3B}]
    \scriptsize
    "The image shows, in the center and slightly tilted to the left, the wreckage of an abandoned military aircraft on dark, flat terrain, probably sandy or volcanic. The fuselage is partially destroyed: the nose is crushed and missing the cockpit, while the tail is visible on the right, with circular openings and signs of corrosion. On the left side of the aircraft, an intense light emanates from the ground, illuminating the gray, oxidized metal structure, which exhibits rough textures and irregular holes. Two bright white lines, resembling lightning trails or photographic effects, extend horizontally from both sides of the aircraft, originating from light sources on the ground. The background consists of a dark blue, uniform, starless night sky, creating a stark contrast with the illuminated elements. The entire scene is plunged into darkness, with the aircraft as the central focal point, surrounded by artificial lights that accentuate its contours and deformations."
    \end{tcolorbox}

    \caption{Qualitative comparison between dense and MoE generations regarding an image from the Incidents1M dataset \cite{Incidents1M}. While the 4B model misclassifies the crashed plane as a boat, the 35B-A3B model accurately recognizes the military aircraft, the structural damage and contextual texts. The generations were translated from the italian language. }
    \label{fig:plane_crash}
\end{figure}

The inference infrastructure was deployed on a server equipped with multiple NVIDIA H100 GPUs. To maximize throughput, the models were physically isolated on separate GPUs and orchestrated via the vLLM framework. The client-side application leveraged asynchronous Input/Output, pushing up to 48 concurrent workers to maintain the server in a state of continuous batching. An asynchronous one-shot warmup mechanism was also implemented to mathematically isolate structural formatting tokens (ChatML) from the KV cache, ensuring precise tracking of the computational cost for each individual image. Examples of generations given by the two architectures are shown in Fig. \ref{fig:plane_crash}.

\vspace{1em}

\subsection{Prompt Configuration for Spatial Objectivity}
When generating captions for Data-Free Knowledge Distillation, conversational and highly creative outputs are detrimental, as they introduce semantic noise into the student model's training targets. To enforce strict objectivity, we constrained the generative temperature to $0.2$, effectively minimizing entropy and reducing the risk of object hallucination \cite{li-etal-2023-evaluating}.

The core semantic constraint was applied through a rigidly structured system prompt. The models were instructed to act as objective disaster response analysts, avoiding any subjective deductions or emotional language. Crucially, the prompt mandated the inclusion of exact spatial coordinates (e.g., "in the foreground", "in the top left corner", "in the background") . This forced spatial mapping functions as an architectural prerequisite because, in a multimodal DFKD framework, encoding specific positional relationships within the text is essential to guide the synthetic generation of spatial features in the surrogate visual data. This ensures proper alignment between the teacher's latent space and the student's representations.

\section{Methodology: Image-Blind LLM-as-a-Judge}
\label{sec:methodology}

Validating 200,000 generated captions via human annotation is economically and temporally unfeasible. Consequently, an automated evaluation protocol is required. In this section, we outline our automated validation framework, detailing why traditional metrics fall short and how an image-blind LLM-as-a-Judge paradigm perfectly simulates the constraints of Data-Free Knowledge Distillation (DFKD).

\subsection{The Bottleneck of Syntactic Metrics}
Historically, automated caption evaluation has relied on n-gram overlap metrics such as BLEU \cite{BLEU}, ROUGE \cite{ROUGE}, or METEOR \cite{METEOR}. However, these syntactic metrics evaluate the surface form of a sentence rather than its semantic meaning. In the critical domain of disaster response, this rigidity introduces two severe failure modes:
\begin{enumerate}
    \item \textbf{Inability to handle logic negations:} Phrases like \textit{``a fire is visible on the building''} and \textit{``no fire is visible on the building''} share nearly identical vocabulary. Syntactic metrics assign an extremely high similarity score to this pair, masking a critical polarity error that would fatally corrupt the student model in DFKD.
    \item \textbf{Insensitivity to synonymy:} Descriptions such as \textit{``a vehicle engulfed in flames''} and \textit{``a burning car''} describe the exact same scene. Since lexical overlap is minimal, standard metrics penalize these paraphrases, failing to recognize their semantic equivalence.
\end{enumerate}
To overcome these bottlenecks, we deployed a dedicated Qwen3.5-9B instance operating as an LLM-as-a-Judge \cite{MTBench, GEval}. A foundational language model can naturally reason over synonyms, negations, and spatial relations, providing a scalable and semantic-aware evaluation.

\subsection{Simulating the Modality Gap: The Image-Blind Paradigm}
A counter-intuitive but foundational design choice of our methodology is that the evaluator LLM (Qwen3.5-9B) is entirely deprived of the original image. The judge operates exclusively on the textual outputs generated by the 4B and 35B models. This \textit{image-blind} approach serves two specific purposes: first, it prevents the judge from forming a third, biased visual interpretation of the scene. If the judge were multimodal, it would inherently compare the generated captions against its own visual perception rather than objectively measuring the textual agreement between the two models. Second, and most importantly, it replicates the operational condition of the Student model during DFKD \cite{FastDKFD}. In a data-free scenario, the Student model never accesses the original training images; it must construct its entire visual understanding derived solely from the text synthesized by the Teacher. By making sure the judge cannot see the image, we measure exactly what the DFKD process requires, which is whether the essential information survived the transition from the visual domain to the textual representation.

\subsection{Task Formulation: Qualitative and Quantitative Assessment}
The evaluation pipeline forces the judge to perform two distinct analytical tasks on the generated dataset:

\textbf{1. Qualitative Task (Semantic Similarity):} To measure the internal agreement between the dense and MoE architectures, the judge receives both Caption A and Caption B simultaneously. We designed a highly constrained prompt instructing the LLM to output a JSON object evaluating four specific criteria on a 0-10 scale: Objects and Subjects, Position and Action, Style and Detail, and Global Similarity. 

\begin{tcolorbox}[title={\itshape Qualitative JSON Prompt Structure}, colframe=black!70, colback=gray!5]
\scriptsize
\textbf{System:} You are an expert judge in semantic and linguistic evaluation of Computer Vision models. Compare Caption A and Caption B generated for the SAME image. \\
\textbf{IMPORTANT:} You will NOT see the image. Your goal is not to determine which model is right, but to analyze the degree of agreement, similarities, and differences. Evaluate based on: \\
1. OBJECTS/SUBJECTS: Do they identify the same main elements? \\
2. POSITION/ACTION: Do spatial relations and actions match? \\
3. STYLE/DETAIL: Is the granularity similar? \\
4. GLOBAL SIMILARITY: Overall semantic agreement (0-10). \\
\textit{Output strictly in JSON format.}
\end{tcolorbox}

\textbf{2. Quantitative Task (Binary Label Validation):} To ground the evaluation against an objective external reference, we validate the generated text against the original Incidents1M multi-label annotations. The judge is prompted iteratively for each associated ground-truth label using a direct binary question: \textit{``Does the following text contain/describe a \{label\}?''}. 
This procedural design (one label, one question) minimizes the LLM's cognitive load. By processing 173,179 unique label pairs, this task explicitly quantifies how many critical disaster features were successfully retained in the textual representations of both the 4B and 35B models.

\section{Results and Analysis}
\label{sec:results}

The automated evaluation pipeline processed 100,000 images, yielding 99,980 valid qualitative comparisons and 173,179 quantitative label verification pairs. The LLM-as-a-Judge formatting success rate exceeded 99.98\%, confirming the extreme stability of the JSON-constrained evaluation prompt.

\subsection{Qualitative Assessment: Semantic Agreement}
\begin{figure}[htbp]
    \centering
    \includegraphics[width=1\linewidth]{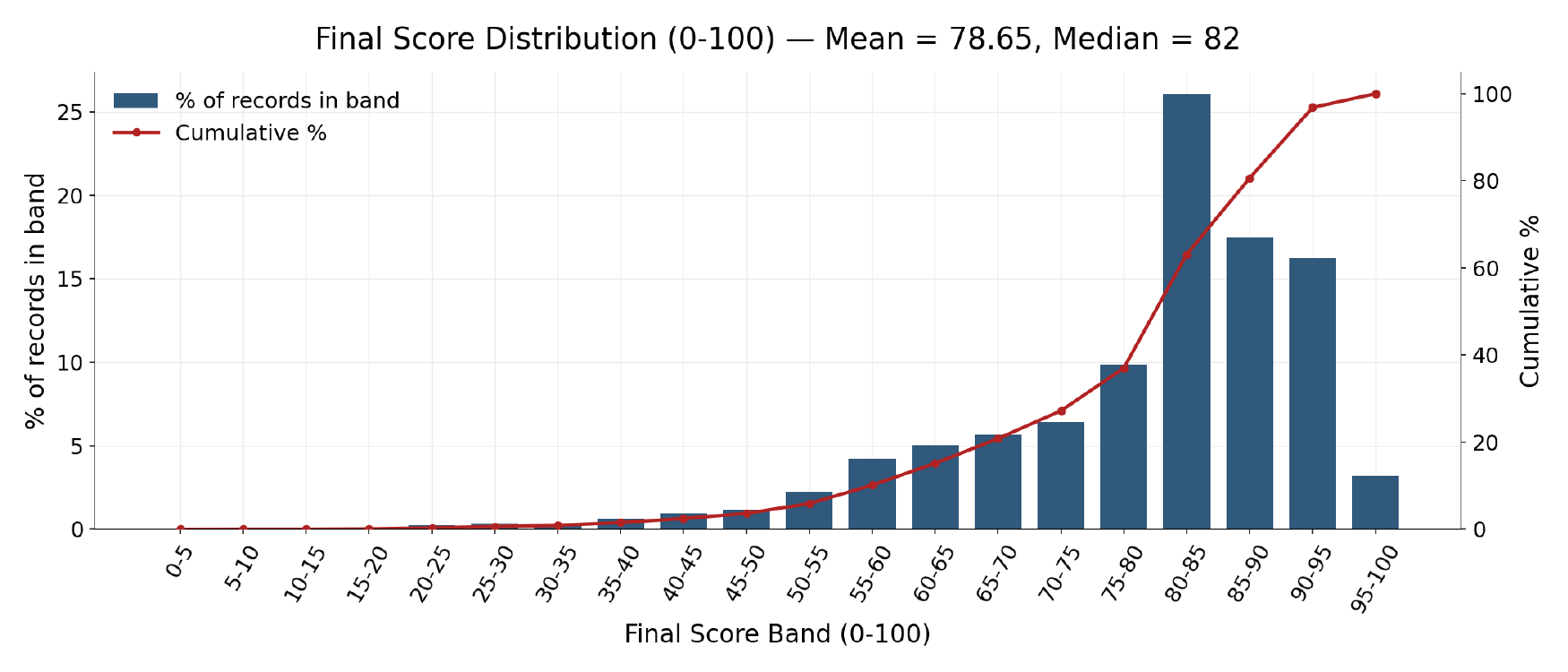}
    \caption{Distribution of the Final Semantic Agreement Score (0-100) between the 4B and 35B-A3B models.}
    \label{fig:score_distribution}
\end{figure}

The qualitative task measured the internal semantic alignment between the dense (4B) and MoE (35B-A3B) architectures without referencing the original ground truth. 
The Final Score, computed as a weighted average of four criteria, reached a mean of 78.65/100 (median 82), with over 59\% of the dataset concentrating in the 80-95 range, as illustrated in Figure \ref{fig:score_distribution}.

Breaking down the criteria reveals where the models converge:
\begin{itemize}
    \item \textbf{Objects and Subjects:} This metric recorded the highest agreement (mean 8.20/10), with 76.1\% of comparisons scoring 8 or above. Both architectures robustly identified the same primary disaster elements (e.g., vehicles, fire, water) regardless of their parameter count.
    \item \textbf{Style and Detail:} This parameter showed the lowest dispersion (standard deviation 0.99) and negligible correlation with the other metrics. This confirms that while the 35B model occasionally employs a richer vocabulary, the stylistic divergence does not alter the fundamental semantic interpretation of the scene.
\end{itemize}

\subsection{Quantitative Assessment: Label Validation for DFKD}
\begin{figure}[htbp]
    \centering
    \includegraphics[width=1\linewidth]{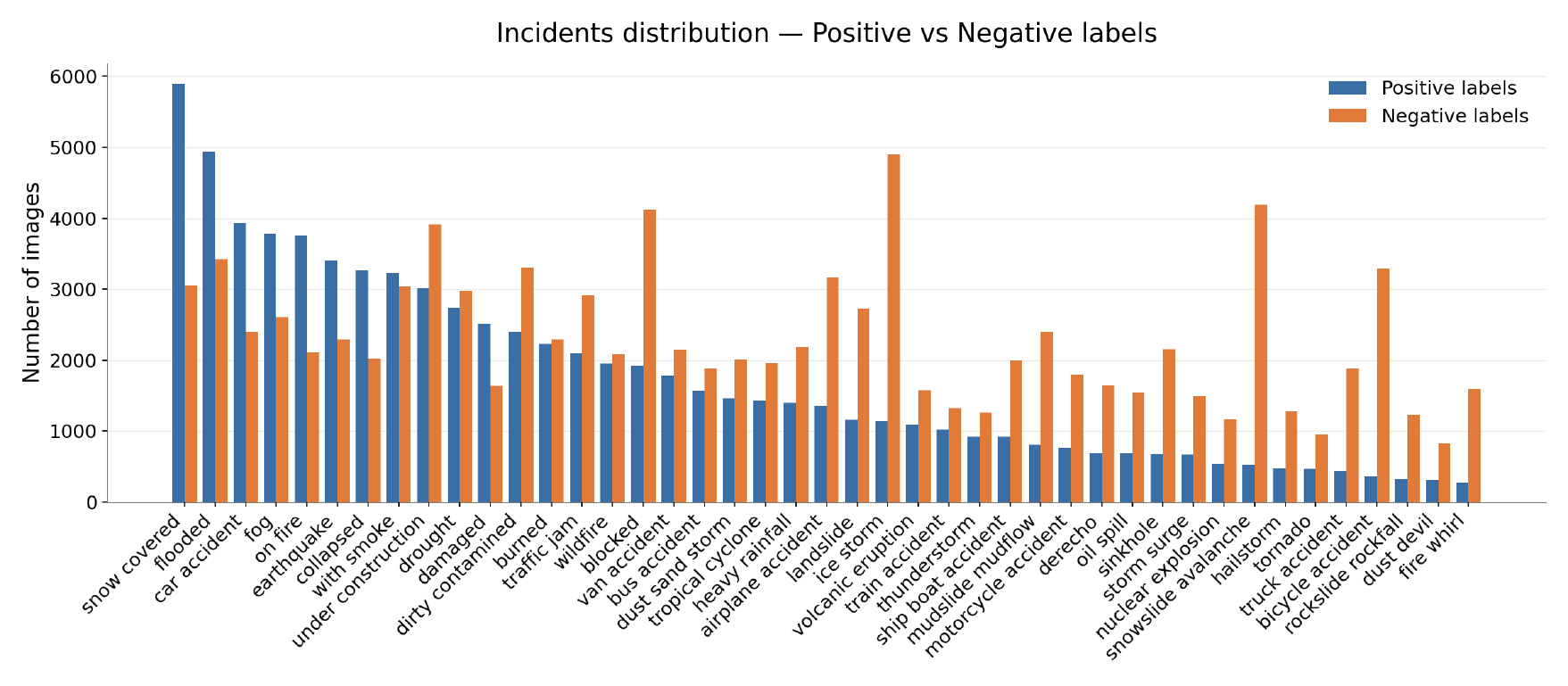}
    \caption{Distribution of positive and negative labels across the 43 incident categories in the evaluated dataset.}
    \label{fig:incidents_distribution}
\end{figure}

The quantitative assessment verified whether the essential disaster features (ground-truth labels) survived the translation from image to text. Table \ref{tab:metrics} aggregates the micro-average performance across all 43 categories. Figure \ref{fig:incidents_distribution} highlights the severe natural imbalance of positive and negative labels across these categories, a critical factor when analyzing aggregate metrics.
\begin{table}[htbp]
\centering
\renewcommand{\arraystretch}{1.35}
\resizebox{\textwidth}{!}{
\begin{tabular}{lccccccccc}
\toprule
\textbf{Model} & \textbf{TP} & \textbf{TN} & \textbf{FP} & \textbf{FN} & \textbf{Precision} & \textbf{Recall} & \textbf{F1-score} & \textbf{Accuracy} & \textbf{Specificity} \\
\midrule
Qwen3.5-4B & 34,205 & 88,969 & 9,859 & 40,140 & 77.6\% & 46.0\% & 57.8\% & 71.1\% & 90.0\% \\
Qwen3.5-35B & 34,836 & 88,555 & 10,275 & 39,511 & 77.2\% & 46.9\% & 58.3\% & 71.3\% & 89.6\% \\
\hline
\end{tabular}
}
\caption{Aggregate micro-average performance of the generated captions evaluated against the 173,179 label pairs of the Incidents1M ground truth.}
\label{tab:metrics}
\end{table}

\begin{figure}[htbp]
    \centering
    \includegraphics[width=1\linewidth]{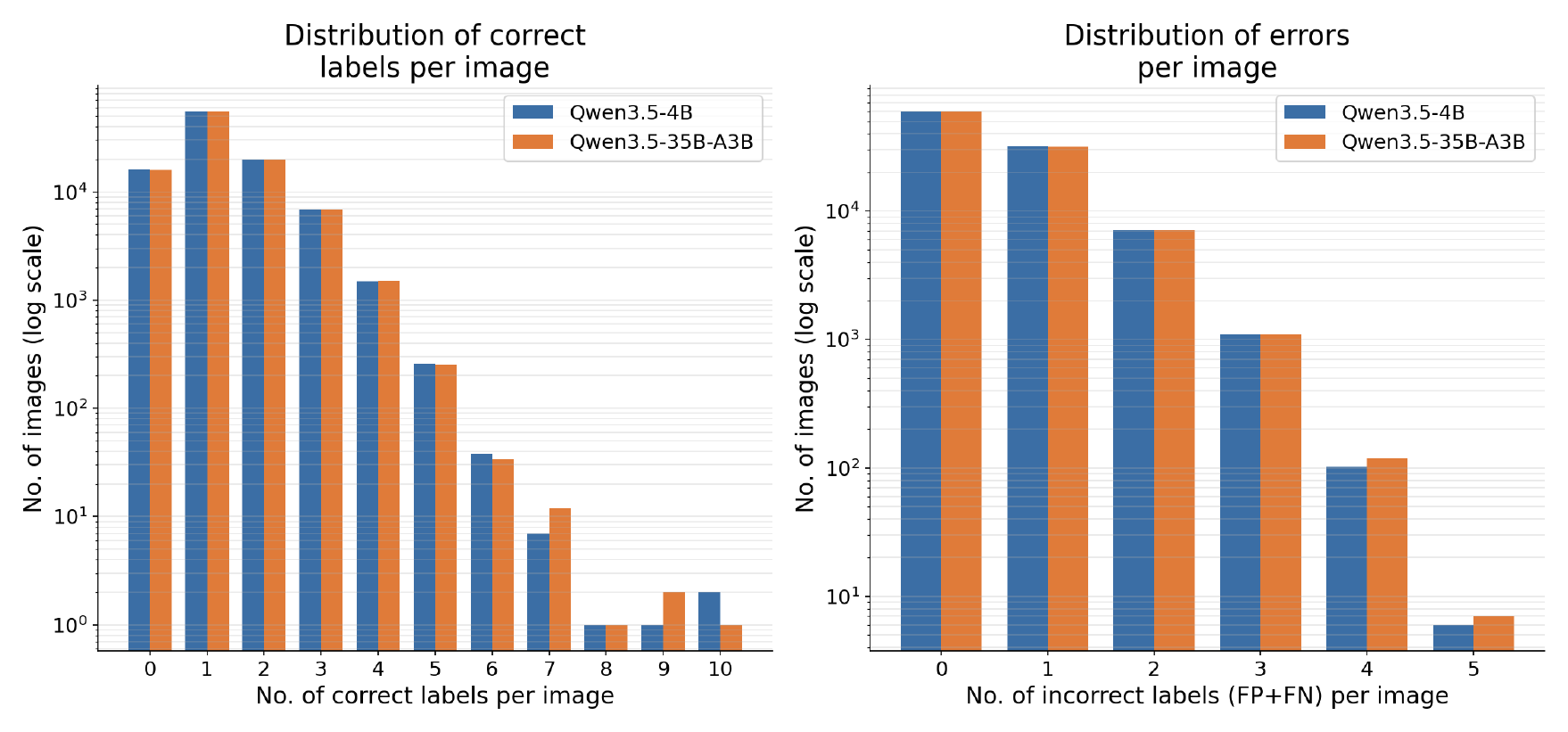}
    \caption{Log-scale distribution of correctly identified labels per image (left) and total errors (False Positives + False Negatives) per image (right).}
    \label{fig:labels_errors}
\end{figure}

A systematic trade-off emerges: both models exhibit high Precision (77.6\% and 77.2\%) but low Recall (46.0\% and 46.9\%). This delineates a highly conservative captioning behavior: the models rarely hallucinate non-existent disasters (Specificity ~90\%), but frequently omit explicit mentions of visually present secondary incidents, as further detailed by the per-image error distributions in Figure \ref{fig:labels_errors}.

In the context of Data-Free Knowledge Distillation, this conservative bias is highly advantageous. A dataset with sparse but highly accurate semantic anchors (low false-positive noise) is strictly preferable to one with exhaustive but hallucinated descriptions, as it prevents the student model from inheriting fabricated cross-modal correlations.

\subsection{Architectural Scaling and the Long Tail of Rare Events}
\begin{figure}[htbp]
    \centering
    \includegraphics[width=1\linewidth]{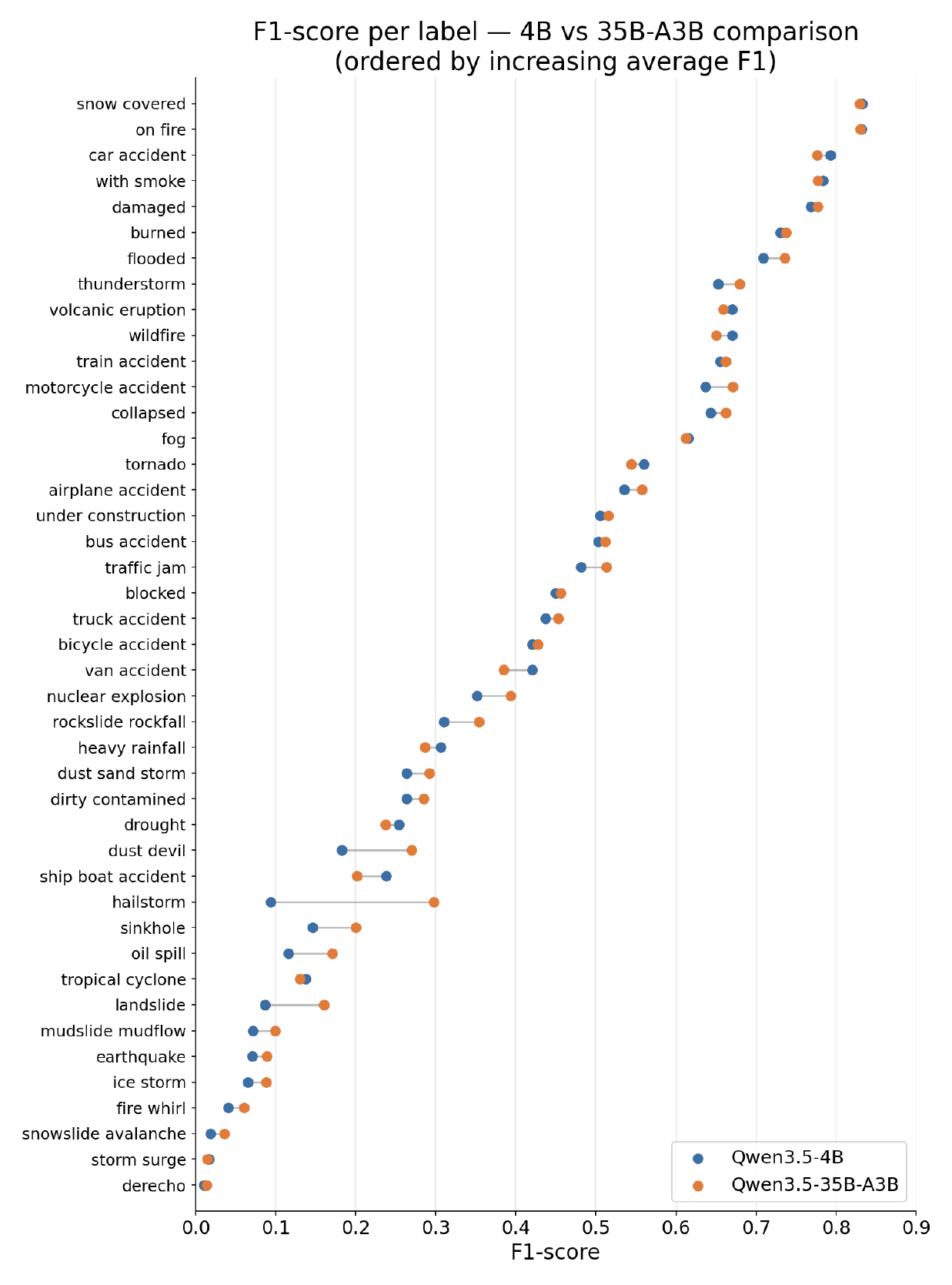}
    \caption{F1-score comparison between the Qwen3.5-4B and 35B-A3B models across the 43 Incidents1M labels, ordered by increasing average performance.}
    \label{fig:f1_dumbbell}
\end{figure}

While the aggregate metrics (Table \ref{tab:metrics}) suggest virtual parity between the 4B and 35B architectures, a macro-average analysis exposes the impact of parametric scaling. Figure \ref{fig:f1_dumbbell} illustrates the F1-score across all 43 categories, revealing a steep gradient. On highly salient categories (e.g., \textit{on fire}, \textit{snow covered}), both models perform identically, achieving F1-scores above 83\%. Conversely, complex meteorological events (e.g., \textit{derecho}, \textit{storm surge}) collapse to near-zero accuracy. This severe degradation suggests that broad, system-level meteorological phenomena (e.g., storm surges) lack the isolated, localized visual anchors that VLMs typically rely on for zero-shot classification.

Table \ref{tab:extremes} explicitly summarizes these extremes, demonstrating how visual prominence dictates textual retrieval. 

\begin{table}[htbp]
\centering
\renewcommand{\arraystretch}{1.2}
\resizebox{\textwidth}{!}{
\begin{tabular}{l | c | c c c c}
\toprule
\textbf{Label} & \textbf{Positive Cases (GT)} & \textbf{F1 4B} & \textbf{F1 35B} & \textbf{Recall 4B} & \textbf{Recall 35B} \\
\midrule
\textbf{on fire} & 3,756 & 83.2\% & 83.0\% & 84.6\% & 84.2\% \\
\textbf{snow covered} & 5,896 & 83.3\% & 83.0\% & 85.6\% & 84.8\% \\
\textbf{car accident} & 3,933 & 79.3\% & 77.6\% & 74.4\% & 71.4\% \\
\textbf{with smoke} & 3,234 & 78.4\% & 77.7\% & 82.6\% & 81.6\% \\
\textbf{damaged} & 2,511 & 76.8\% & 77.7\% & 77.2\% & 79.2\% \\
\midrule
\textbf{tropical cyclone} & 1,427 & 13.8\% & 13.1\% & 8.3\% & 7.8\% \\
\textbf{mudslide mudflow} & 812 & 7.1\% & 9.9\% & 3.8\% & 5.4\% \\
\textbf{earthquake} & 3,404 & 7.1\% & 8.9\% & 3.7\% & 4.7\% \\
\textbf{snowslide avalanche}& 526 & 1.8\% & 3.6\% & 1.0\% & 1.9\% \\
\textbf{storm surge} & 671 & 1.7\% & 1.4\% & 0.9\% & 0.7\% \\
\textbf{derecho} & 691 & 1.1\% & 1.4\% & 0.6\% & 0.7\% \\
\bottomrule
\end{tabular}
}
\caption{Performance extremes for both models. Visually distinct categories achieve high accuracy, while broad contextual disasters suffer from severe recall degradation.}
\label{tab:extremes}
\end{table}

\begin{figure}[htbp]
    \centering
    \includegraphics[width=1\linewidth]{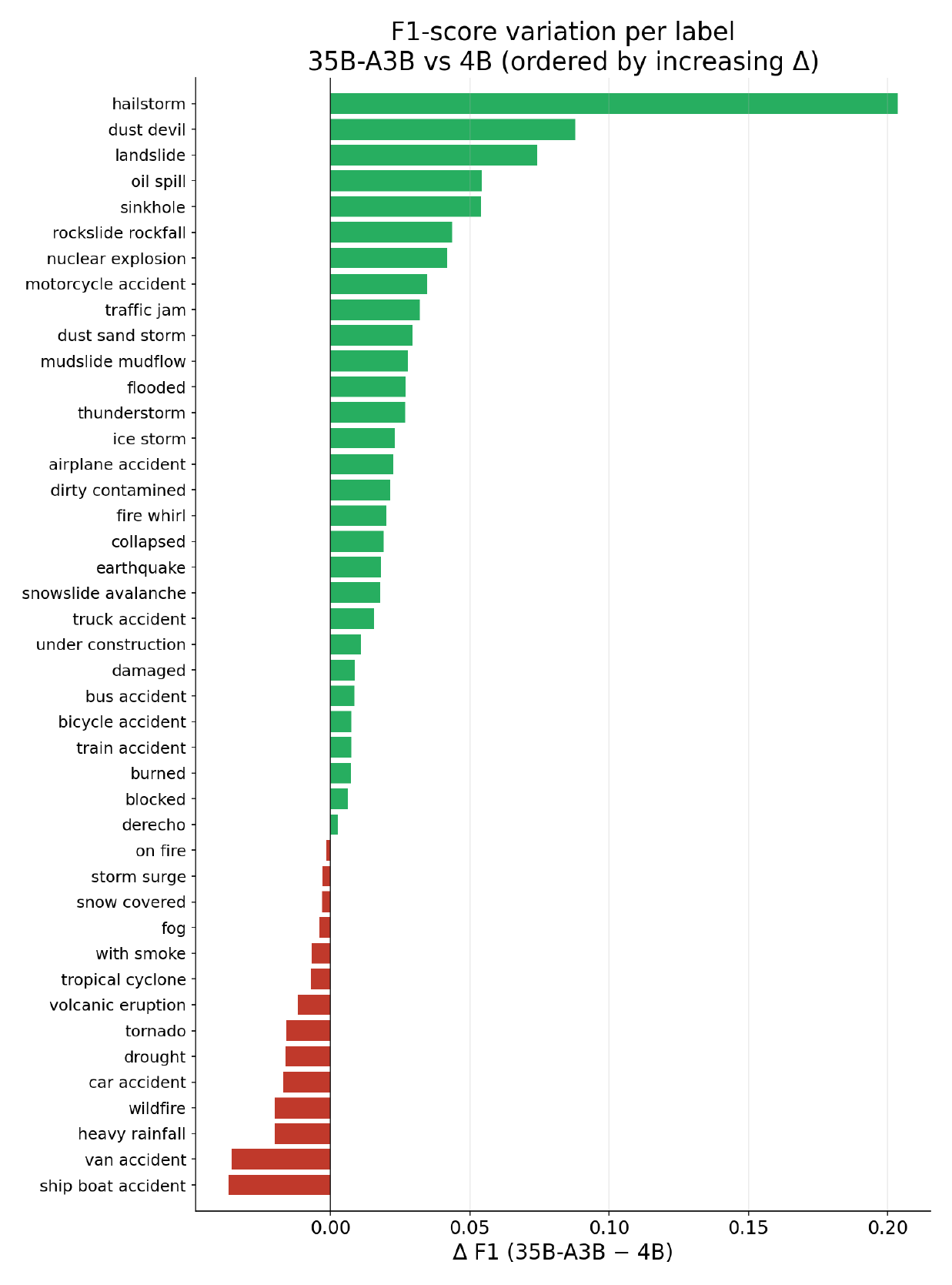}
    \caption{Variation of the F1-score per label (35B-A3B minus 4B). The MoE architecture exhibits a systematic advantage specifically on the long tail of rare categories.}
    \label{fig:delta_f1}
\end{figure}

The primary advantage of the Mixture-of-Experts architecture emerges strictly within the long tail of rare, semantically specific categories. As shown in Figure \ref{fig:delta_f1}, the 35B-A3B model demonstrates a systematic improvement over the 4B model precisely where lexical precision is crucial. The most prominent gains are observed in classes like \textit{hailstorm} (+20.3\% F1), \textit{dust devil} (+8.8\%), and \textit{landslide} (+7.5\%). This confirms that the broader knowledge base accessed via sparse routing provides the advanced technical vocabulary required to successfully caption rare disaster events, without disrupting the baseline perception of common scenarios.

\newpage
\section{Discussion: Ground Truth Inconsistencies}
\label{sec:discussion}

A fundamental assumption in supervised evaluation is that the original dataset annotations, the ground truth, serve as a flawless reference. However, during a manual inspection of the divergent cases where our LLM-as-a-Judge pipeline registered False Positives or False Negatives against the Incidents1M labels, a critical phenomenon emerged: numerous discrepancies were not caused by VLM hallucinations, but by intrinsic inconsistencies in the original human annotations.

\subsection{Human Annotation Fallacies}
The original Incidents1M dataset was labeled via crowdsourcing (Amazon Mechanical Turk), where operators were asked binary questions about the presence of specific incidents. Our analysis revealed two distinct typologies of human error that actively corrupted the evaluation metrics.

\begin{figure}[htbp]
    \centering
    \begin{subfigure}{0.48\textwidth}
        \centering
        \includegraphics[width=\linewidth]{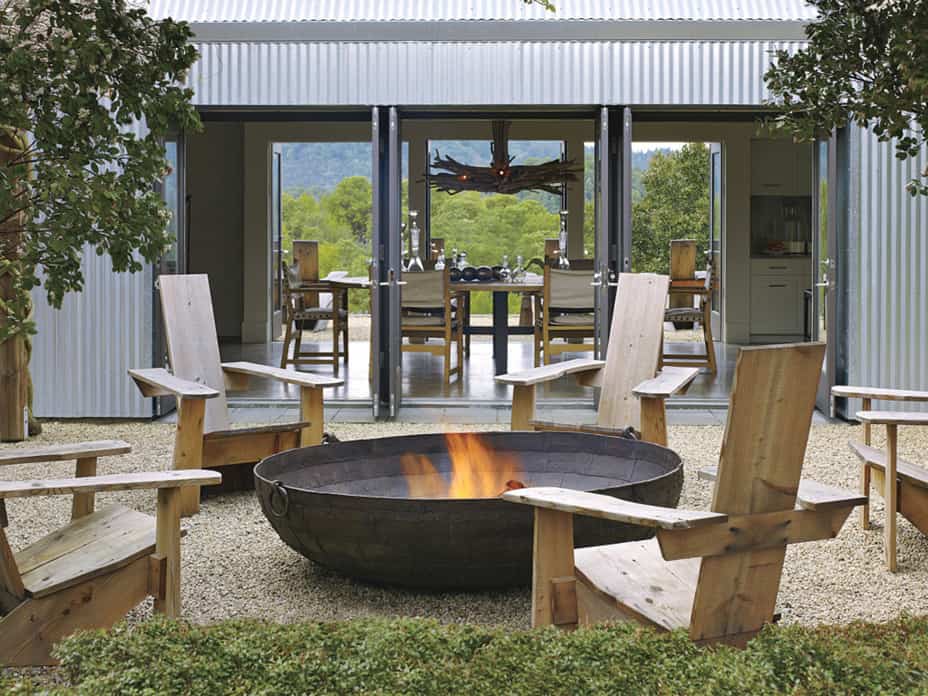}
        \caption{Ground Truth: \textit{on fire = False}}
    \end{subfigure}\hfill
    \begin{subfigure}{0.48\textwidth}
        \centering
        \includegraphics[width=\linewidth]{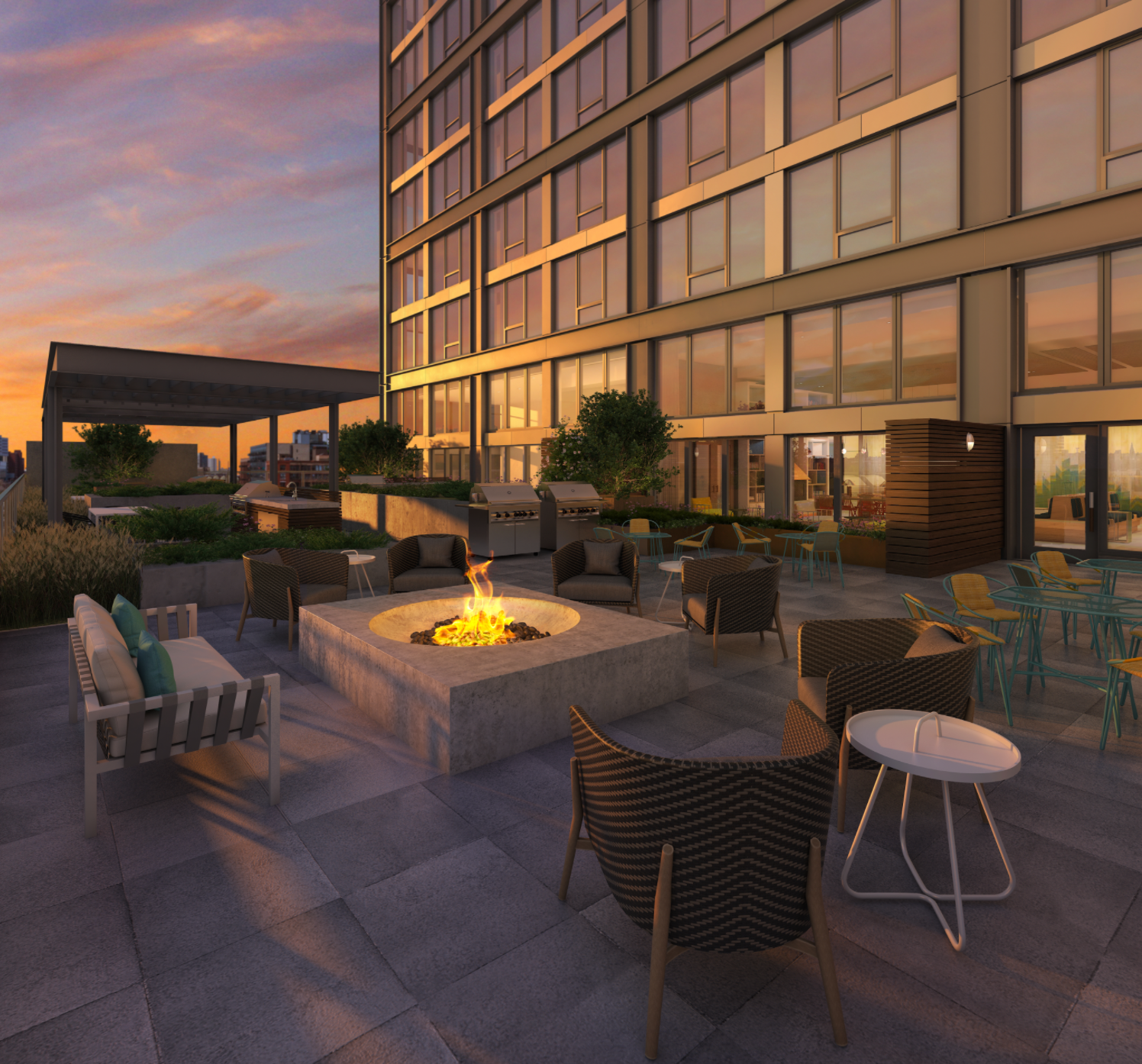}
        \caption{Ground Truth: \textit{on fire = True}}
    \end{subfigure}
    \caption{Inconsistent ground truth labeling in Incidents1M \cite{Incidents1M}. Both images depict outdoor contained fires, yet they received contradictory human annotations.}
    \label{fig:fire_inconsistency}
\end{figure}

First, we observed a severe lack of consistency in defining ambiguous scenarios. For example, regarding the \textit{on fire} category, visually nearly identical images depicting small, contained fires (such as outdoor braziers or campfires) were labeled oppositely by human annotators. As shown in Figure \ref{fig:fire_inconsistency}, one image was labeled as \textit{True} while the other as \textit{False}, despite the lack of any objective visual criterion justifying the discrepancy.

\begin{figure}[htbp]
    \centering
    \begin{subfigure}{0.48\textwidth}
        \centering
        \includegraphics[width=\linewidth]{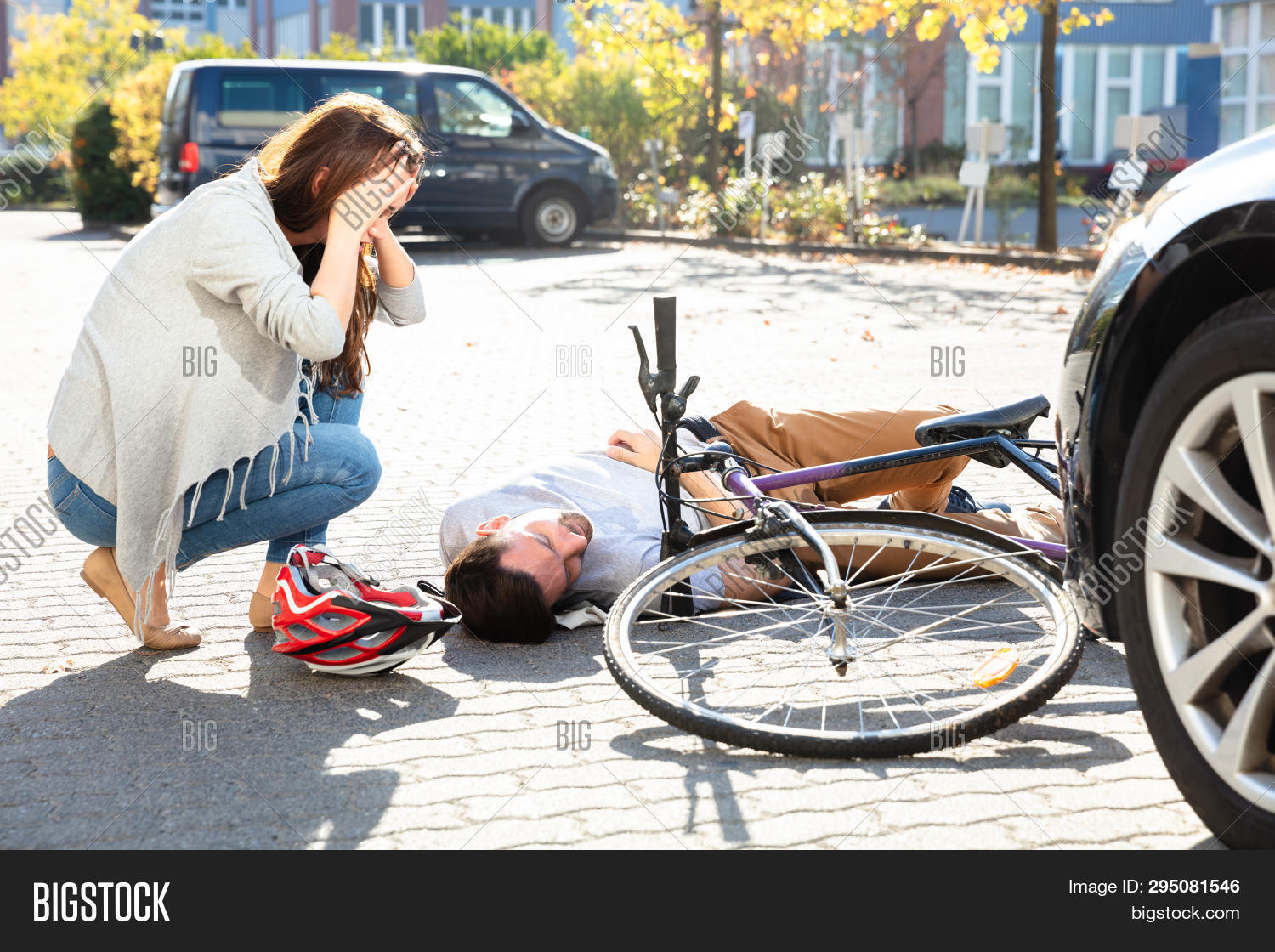}
        \caption{Missing label: \textit{bicycle accident}}
    \end{subfigure}\hfill
    \begin{subfigure}{0.48\textwidth}
        \centering
        \includegraphics[width=\linewidth]{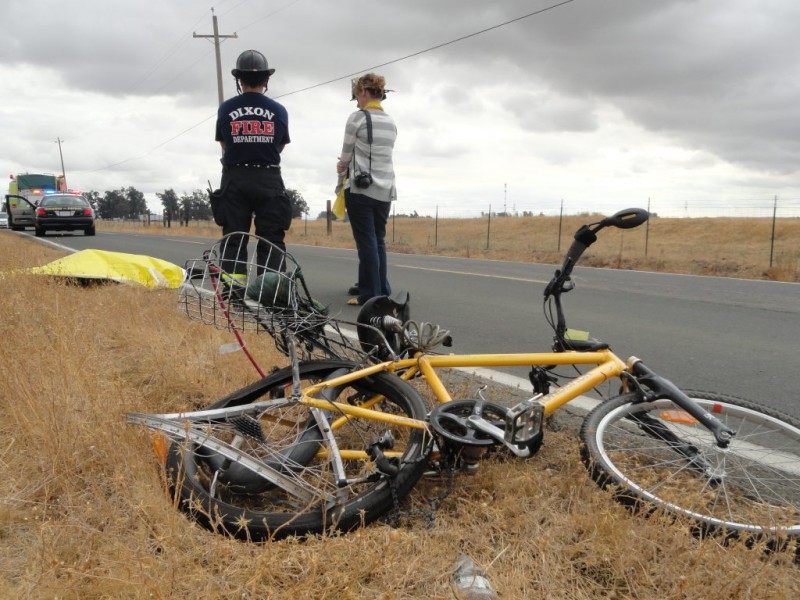}
        \caption{Missing label: \textit{bicycle accident}}
    \end{subfigure}
    \caption{Ground truth omissions. In both images, the VLMs correctly described severe bicycle accidents. However, since the human annotators missed these labels, the automated evaluation unfairly penalized the generative models with False Positives.}
    \label{fig:bike_omission}
\end{figure}

Second, we found instances of severe omission (human False Negatives). In several cases, the Qwen teacher models correctly identified and extensively detailed incidents, such as a bicycle accident, that were clearly visible in the imagery but completely absent from the original positive labels (see Figure \ref{fig:bike_omission}). When our automated judge evaluated the accurate generated text against this incomplete ground truth, it inevitably registered a False Positive. Consequently, the generative models were mathematically penalized for accurately describing a scene that the human annotators had missed.

\subsection{Implications for Future Benchmarks}
This discovery carries a methodological implication. The Precision, Recall, and F1-score metrics reported in Section \ref{sec:results} must be interpreted as a lower bound of the VLMs' true descriptive capabilities. A non-quantifiable portion of the recorded errors derives directly from ground truth inaccuracies rather than genuine generative flaws. 

Ultimately, this demonstrates that modern, highly capable VLMs possess a level of zero-shot visual reasoning that rivals, and sometimes exceeds, the reliability of crowdsourced human workers. Moving forward, robust Vision-Language Models should be systematically employed not merely to train on or consume existing multimodal datasets, but to actively audit, clean, and refine large-scale human-annotated benchmarks before they are used for downstream tasks or Knowledge Distillation.

\section{Conclusion and Future Work}
\label{sec:conclusion}

In this work, we addressed the critical shortage of high-quality multimodal datasets for disaster response, a mandatory prerequisite for Data-Free Knowledge Distillation (DFKD). By engineering an atomic dataset reconstruction pipeline, we successfully recovered 100,000 images from the vision-only Incidents1M dataset and enriched them with 200,000 highly objective, spatially-aware textual captions using two distinct Qwen3.5 architectures.

To overcome the fragility of traditional syntactic metrics and strictly simulate the modality gap inherent to DFKD, we introduced an innovative image-blind LLM-as-a-Judge validation framework. Our extensive evaluation demonstrated a massive semantic agreement (78.65/100) between the 4B dense and 35B Mixture-of-Experts models. While both architectures exhibit a conservative generative behavior that minimizes fatal false-positive hallucinations, our analysis revealed that the sparse MoE paradigm provides a decisive advantage in captioning the long tail of rare and semantically complex disaster events.

Crucially, our automated evaluation enabled us to discover underlying human-annotation fallacies within the original ground truth, highlighting the capability of modern VLMs to act as rigorous benchmark auditors. The resulting dataset and the reproducible validation methodology presented in this paper establish a clean, LLM-validated semantic foundation to advance cross-modal knowledge distillation in mission-critical environments.

Building upon these findings, our future work will focus on executing the end-to-end Data-Free Knowledge Distillation pipeline. We aim to leverage this validated multimodal dataset to train compact student networks capable of real-time visual disaster classification on resource-constrained edge devices, such as search-and-rescue drones. Furthermore, expanding our VLM-driven auditing framework to systematically clean, re-annotate, and expand the entirety of the Incidents1M benchmark represents a critical next step toward providing the community with a robust multimodal resource.
\newpage
\bibliographystyle{elsarticle-num} 
\bibliography{bibliography}
\end{document}